\documentclass[a4paper,11pt]{article}

\usepackage{times}
\usepackage{graphicx}
\usepackage{amsmath}
\usepackage[usenames]{color}
\usepackage{booktabs}
\usepackage{caption}
\usepackage{subfig}
\usepackage{transparent}
\usepackage{tabularx}
\usepackage[ruled,lined]{algorithm2e}

\definecolor{cwblue1}{rgb}{0.27,0.427,0.623}
\definecolor{cwblue2}{rgb}{0.286,0.454,0.658}
\definecolor{cwblue3}{rgb}{0.733,0.811,0.905}

\newcommand{\hypot}[1]{
{\bf Assumption #1} ---
}

\newcommand{\change}[1]{#1}

\newcommand{\overm}{\overline{M}}
\newcommand{\overs}{\overline{S}}

\newcommand{\etal}{{\em et al. }}

\newcommand{\executeiffilenewer}[3]{%
\ifnum\pdfstrcmp{\pdffilemoddate{#1}}%
{\pdffilemoddate{#2}}>0%
{\immediate\write18{#3}}\fi%
}

\begin{document}
%
\title{Activity recognition from videos with parallel hypergraph matching on GPUs}

\author{Eric Lombardi$^1$  \hspace{5mm} 
	Christian Wolf$^{1,2}$  \hspace{5mm}
	Oya \c{C}eliktutan$^3$  \hspace{5mm}
	B\"{u}lent Sankur$^3$ 
\\
\\
    $^1$ Universit\'{e} de Lyon, CNRS, LIRIS UMR 5205, France \\		   
	$^2$ INSA-Lyon, F-69621, France \\
	$^3$ Bo\u{g}azi\c{c}i University, Dept. of Electrical-Electronics Eng., Turkey \\
}

\maketitle

\begin{abstract}
In this paper, we propose a method for activity recognition from videos based on sparse local features and hypergraph matching. We benefit from special properties of the temporal domain in the data to derive a sequential and fast graph matching algorithm for GPUs. 

Traditionally, graphs and hypergraphs are frequently used to recognize complex and often non-rigid patterns in computer vision, either through graph matching or point-set matching with graphs. Most formulations resort to the minimization of a difficult discrete energy function mixing geometric or structural terms with data attached terms involving appearance features. Traditional methods solve this minimization problem approximately, for instance with spectral techniques. 

In this work, instead of solving the problem approximatively, the exact solution for the optimal assignment is calculated in parallel on GPUs. The graphical structure is simplified and regularized, which allows to derive an efficient recursive minimization algorithm. The algorithm distributes subproblems over the calculation units of a GPU, which solves them in parallel, allowing the system to run faster than real-time on medium-end GPUs.
\end{abstract}


\section{Introduction}
\label{sec:introduction}
Many computer vision problems can be formulated as graphs and associated algorithms, since graphs provide a structured and flexible way to inject spatial and structural relationships into matching algorithms. In this paper, it is employed for recognition and localization of actions in videos. The task to detect and localize activities in time and in space and to classify them requires dealing with large quantities of video data in real time.

We formulate the activity recognition task as a correspondence problem between sparse features of short duration model actions and those of longer duration scene videos. The articulated nature of human motions makes it impossible to employ rigid transformations and methods (like RANSAC \cite{FischlerRANSAC1981}), while graph matching is able to cope with such non-rigid deformations. The proposed method structures space-time interest points into graphs or hypergraphs using proximity information, as is frequently done in the context of visual recognition (object or activity recognition). The optimal matching between a model video and a scene video is cast as a combinatorial problem to minimize a function containing terms measuring differences in appearance features as well as terms adressing space-time geometric distortions in the matching. 

Despite advances in graph matching due to its popularity and effectiveness, it still remains a challenging task. The computational complexity renders the use of the exact problem on data having a large number of nodes, e.g. video data, intractable in practice. Formulations useful in vision are known to be NP-hard: while the graph isomorphism problem is conjectured to be solvable in polynomial time, it is known that exact subgraph matching is NP-hard \cite{Torresani2008ECCV}, and so is subgraph isomorphism \cite{GareyJohnson1979}. Graph matching solutions in practical problems therefore are of approximate nature.

Recently, computer vision has immensely benefited from development of general purpose computation on graphical processing units (GPUs). Prominent examples are classification in various applications (e.g. pose estimation \cite{Shotton_depthpose_CVPR11}) and convolution operations, for instance in deep learning architectures \cite{Krizhevsky2012}, feature tracking and matching \cite{SinhaFrahm2006} and patch based image operations, as for instance inpainting \cite{WangXiongYun2013}. While it is straightforward to profit from parallel architectures if the problem is inherently parallelizable and data-oriented, structured problems are often characterized by complex dependencies which make parallelization of the algorithms difficult. The main inpediments are irregular memory access, tree-based search structures, variable computation and memory units requirement \cite{Jenkins2011}.

Whereas most existing work solves the matching problem approximatively, in this work the exact global minimum is calculated, which is made possible by two properties: 
\begin{itemize}
\item We benefit from two sources, first from the application itself (activity recognition in videos) and the fact that the data are embedded in space time, in particular from specific properties of the time dimension;
\item We approximate the graphical structure and therefore solve a simplified problem exactly and efficiently. The work thus falls into the category of methods calculating the exact solution for an approximated model, unlike methods calculating an approximate solution of an exact problem. In this sense, it can be compared to \cite{CaetanoCaelliBarone2006}, where the original graph of a 2D object is replaced by a k-tree allowing exact minimization by the junction tree algorithm. Our solution is different in that the graphical structure is not created randomly but is derived from the temporal dimensions of video data. 
\end{itemize}
The linearly ordered nature of actions in time allows matching to proceed through a recursive algorithm over time, which is sequential in nature. However, the subproblems corresponding to calculations associated for individual time instants involve iterations over the domains of discrete variables, which are independent. Faster than real-time performance is achieved through a parallel solution of these subproblems on GPUs.

\subsection{Related work}


\noindent
\textbf{Graphs and graph matching in computer vision} ---
In computer vision, graph matching has been mainly applied for object recognition, but some applications to action recognition problems have recently been reported. In this context, graphs are frequently constructed from sparse primitives like space time interest points \cite{Ta_grapmatching_AVSS10,Gaur_StringFeatureGraphs_ICCV11,Borzeshi_PrototypeSelection_ICCVW11} or from regions and adjacency information gathered from over-segmented videos \cite{BrendelTodorovic2011}. Common matching strategies in this context are off-the-shelf spectral techniques \cite{Ta_grapmatching_AVSS10} or dynamic time warping on strings of graphs \cite{Gaur_StringFeatureGraphs_ICCV11}.

Other graph related work, albeit not necessarily by matching, is based on stochastic Kronecker graphs \cite{TodorovicECCV12} for modeling the features pertaining to a specific activity class, chain-graphs \cite{ZhangZengJiITIP11}, weighted directed graphs \cite{LiZhangLiuCSVT08} where nodes are assigned to frames with salient postures or silhouettes that are shared among different action categories. Here, edges encode the transition probabilities between these postures. Salient postures are used for single-view and multi-view action recognition in \cite{LvNevatiaCVPR07}. In \cite{LvNevatiaCVPR07}, different actions are linked by chain graphs, called as Action Nets. In \cite{Chen_MaxSubgraph_CVPR12}, the test video is first divided into several subvolumes and the maximum subgraph is searched through branch-and-cut algorithms.

Graph matching and graphs are not the only technique successfully employed for activity recognition. A review of the state of the art of this problem is beyond the scope of this paper, we refer the reader to \cite{Aggarwal_review_ACM11} for a recent survey. 

\begin{figure*}[t] \centering
\begin{tabular}{cc}
	\begin{minipage}{5cm}
	\includegraphics[width=5cm]{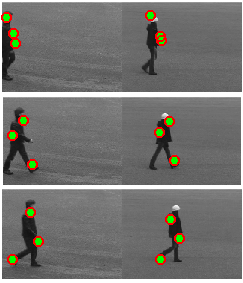}
	\end{minipage} 
	&
	\begin{minipage}{9cm}
	\includegraphics[width=9cm]{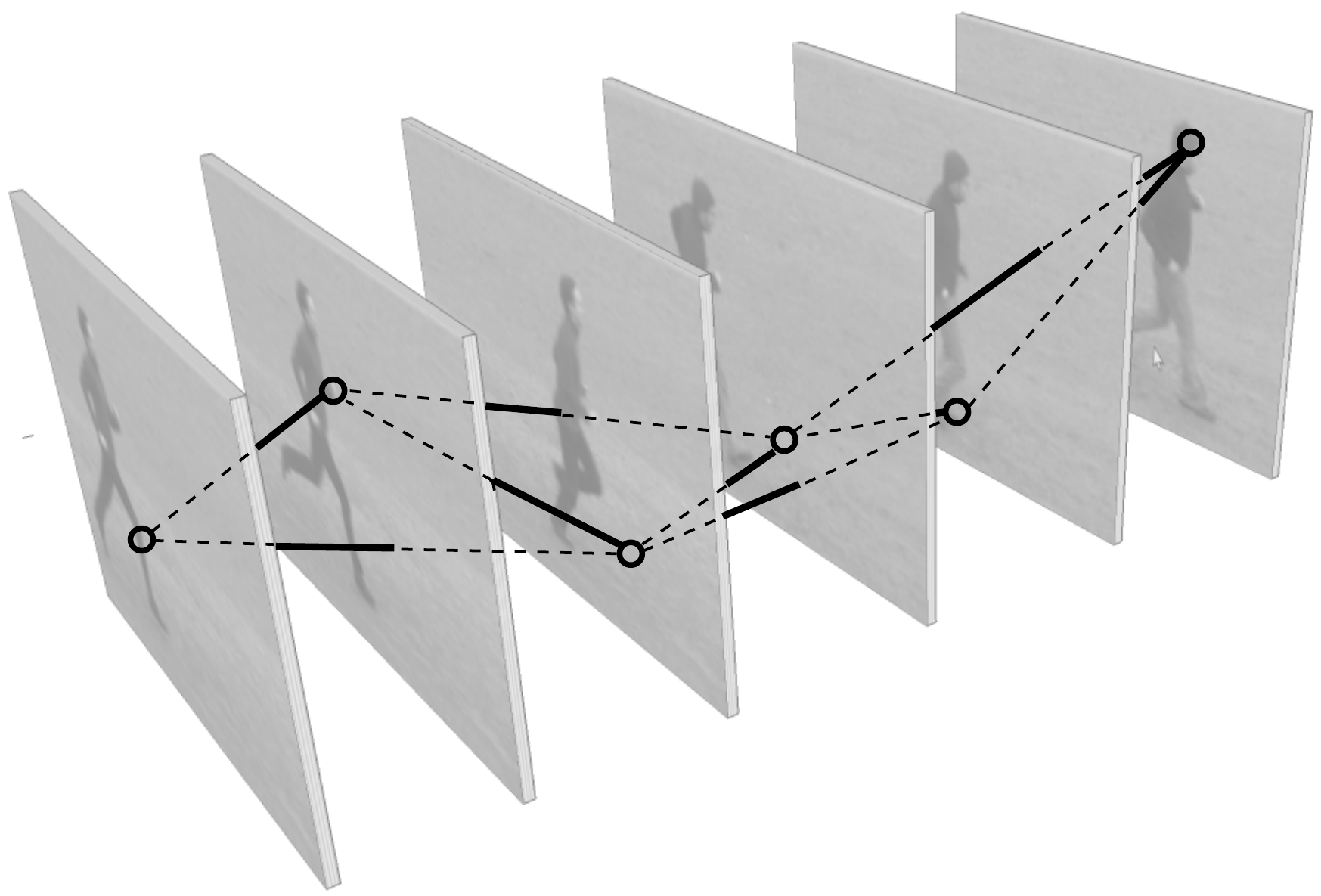}
	\end{minipage}
	\\ (a) & (b) \\ 	
\end{tabular}
\caption{Videos are represented through space time interest points: (a) examples of successful matches of space time points from the model video (left) to the scene video (right); (b) model videos are structured into graphs with a specific structure using proximity information.\label{fig:stipsgraph}}
\end{figure*}


\textbf{Solving the graph matching problem} ---
Two different formulations dominate the literature on graph matching: 
(i) Exact matching: a strictly structure-preserving correspondence between the two graphs or at least between their respective parts is searched; (ii) Inexact matching, where compromises in the correspondence are allowed in principle by admitting structural deformations up to some extent. Matching proceeds by minimizing an objective (energy) function.  

Most recent papers on graph matching in the computer vision context are based on inexact matching of valued graphs, i.e., graphs with additional geometric and/or appearance information associated with nodes and/or edges. Practical formulations of this problem are known to be NP-hard \cite{Torresani2008ECCV}, which makes approximations unavoidable. Two different strategies are frequently used: 
(i) Calculation of an approximate solution  of the minimization problem;
(ii) Calculation of the exact solution, most frequently of an approximated model.

{\it Approximate solution ---} A well known family of methods solve a continuous relaxation of the original combinatorial problem. Zass and Shashua \cite{ZassCVPR2008} presented a soft hypergraph matching method between sets of features that proceeds through an iterative successive projection algorithm in a probabilistic setting. They extended the Sinkhorn algorithm \cite{Knight06thesinkhorn-knopp}, which is used for soft assignment in combinatorial problems, to obtain a global optimum in the special case when the two graphs have the same number of vertices and an exact matching is desired. They also presented a sampling scheme to handle the combinatorial explosion due to the degree of hypergraphs. Zaslavskiy \etal \cite{ZaslavskiyBachVert2009} employed a convex-concave programming approach to solve the least-squares problem over the permutation matrices. 
More explicitly, they proposed two relaxations to the quadratic assignment problem over the set of permutation matrices which results in one quadratic convex and one quadratic concave optimization problem. They obtained an approximate solution of the matching problem through a path following algorithm that tracks a path of local minimum by linearly interpolating convex and concave formulations. 

A specific form of relaxation is done by spectral methods, which study the similarities between the eigen-structures of the adjacency or Laplacian matrices of the graphs or of the assignment matrices corresponding to the minimization problem formulated in matrix form. In particular, Duchenne \etal \cite{DuchenneCVPR09} generalized the spectral matching method from the pairwise graphs presented in \cite{LeordeanuICCV2005} to hypergraphs by using a tensor-based algorithm to represent affinity between feature tuples, which is then solved as an eigen-problem on the assignment matrix. More explicitly, they solved the relaxed problem by using a multi-dimensional power iteration method, and obtained a sparse output by taking into account $l_1$-norm constraints instead of the classical $l_2$-norm. 
Leordeanu \etal \cite{LeordeanuZanfirSminchisescu2011} made an improvement on the solution to the integer quadratic programming problem in \cite{DuchenneCVPR09} by introducing a semi-supervised learning approach. In the same vein, Lee \etal \cite{LeeChoLee2011} approached this problem via the random walk concept.

Another approach is to decompose the original discrete matching problem into subproblems, which are then solved with different optimization tools. A case in point, Torresani \etal \cite{Torresani2008ECCV} solved the subproblems through graph-cuts, Hungarian algorithm and local search. Lin \etal \cite{LinZengLiuZhu2009} first determined a number of subproblems where each one is characterized by local assignment candidates, i.e., by plausible matches between model and scene local structures. For example, in action recognition domain, these local structures can correspond to human body parts. Then, they built a candidacy graph representation by taking into account these candidates on a layered (hierarchical) structure and formulated the matching problem as a multiple coloring problem. Finally, Duchenne \etal \cite{DuchenneJoulinPonce2011} extended one dimensional multi-label graph cuts minimization algorithm to images for optimizing the Markov Random Fields (MRFs). 

\noindent
{\it Approximate graphical structure ---} An alternative approach is to approximate the data model, for instance the graphical structure, as opposed to applying an approximate matching algorithm to the complete data model. One way is to simplify the graph by filtering out the unfruitful portion of the data before matching. For example, a method for object recognition has been proposed by Caetano \etal \cite{CaetanoCaelliBarone2006}, which approximates the model graph by building a k-tree randomly from the spatial interest points of the object. Then, matching was calculated using the classical junction tree algorithm \cite{LauritzenSpiegelhalter1988} known for solving the inference problem in Bayesian Networks.

A special case is the work by Bergthold et al. \cite{Bergtholdt2010}, who perform object recognition using fully connected graphs of small size (between 5 and 15 nodes). The graphs can be small because the nodes correspond to semantically meaningful parts in an object, for instance landmarks in a face or body parts in human detection. A spanning tree is calculated on the graph, and from this tree a graph is constructed describing the complete state space. The $A^*$ algorithm then searches the shortest path in this graph using a search heuristic. The method is approximative in principle, as hypotheses are discarded due to memory requirements. However, for some of the smaller graphs used in certain applications, the exact solution can be obtained.


\textbf{Parallel graph matching} ---
Parallel algorithms have been proposed for the graph matching problem for some time. Although both exact solutions and approximate solutions, can be parallized in principle, most of the existing parallel algorithms have been proposed for approximate solution. Many spectral methods, which are approximative, can be naturally ported to a GPU architecture, as the underlying numerical algorithms require matrix operations. Similar matrix operations are employed in \cite{RodenasFrencesc2011}, where multiple graphs are matched using graduated assignment on GPUs.

Matching two graphs can also be performed by searching the maximum common subgraph of the two graphs \cite{Conte_challengingcomplexity2007}, a problem which can be transformed (in linear time) to the problem of searching the maximum clique of a graph. For this problem parallel and GPU algorithms do exist, albeit not very efficient ones \cite{Jenkins2011}.

Parallel and GPU algorithms for bi-partite matching have been proposed recently \cite{VasconcelosRosenhahn2009,Kollias2012}. Bi-partite matching is, however, different and less difficult, as polynomial time algorithms are known for them. These algorithms alternate bidding (proposed assignments) kernels and assignment kernels on the GPU as well as convergence tests on the CPU.   A similar problem, unfortunately also called graph matching, matches neighboring vertices of a single graph under unicity constraints. In other terms, a \textit{matching} or independent edge set in a graph is a set of edges without common vertices. GPU algorithms have been proposed for this kind of problem \cite{Faggingerauer2012}.


\section{Problem Formulation}
\label{sec:problem}
\noindent
Detecting, recognizing and localizing activities in a video stream is cast as a matching problem between a scene video, which can be potentially long or short if the data is processed block-wise in a stream, and a dictionary of relatively much shorter model videos describing the set of activities to recognize. Matching is done by pairs, solving a correspondence problem between the scene video and a single model video at a time. We formulate the problem as a particular case of the general correspondence problem between two point sets with the objective of assigning points from the model set to points in the scene set, such that some geometrical invariance is satisfied. In particular, videos are represented as space-time interest points --- see Figure \ref{fig:stipsgraph}a for examples of successful matchings between model point sets and scene point sets. Each point is described by its location in space-time and appearance features, i.e., a descriptor locally representing the space-time region around the point.

The $M$ points of the model are organized as a hypergraph $\mathcal{G}{=}\{\mathcal{V},\mathcal{E}\}$, where $\mathcal{V}$ is the set of nodes (corresponding to the points) and $\mathcal{E}$ is the set of edges. Let us recall that hypergraphs are a generalization of graphs, where edges, often called \textit{hyperedges}, can link any number of nodes, generally ${>}2$. The set of scene nodes is not structured. 

Each node $i$ of the model graph is assigned a discrete variable $z_i,\ i=1..M$, which represents the mapping from the $i^{th}$ model node to some scene node, and can take values in $1..S$, where $S$ is the number of scene nodes. We use the shorthand notation $z$ to denote the whole set of map variables $z_i$. A solution of the matching problem is given through the values of the $z_i$, where $z_i{=}j$, $i{=}1..M$, is interpreted as model node $i$ being assigned to scene node $j=1..S$. Each combination of assignments $z$ evaluates to an energy value in terms of the following energy function $E(z)$:

\begin{equation}
E(z) = \lambda_1 \sum_i U(z_i) \ + \ \lambda_2 \sum_{(i,j,k)\in\mathcal{E}} D(z_i,z_j,z_k)
\label{eq:globalenergyhypershort}
\end{equation}

\change{
\noindent
Here, $U$ is a data attached term taking into account the distance between appearance features of point $i$ and its assigned point $z_i$,  $D$ is the geometric distortion between the space-time triangle associated with hyperedge $(i,j,k)$ and the triangle associated with $(z_i,z_j,z_k)$, and $\lambda_1$ and $\lambda_2$ are weights. For convenience, dependencies on all values over which we do not optimize have been omitted from the notation.

$U$ is defined as the Euclidean distance between the appearance features of assigned points in the case of a candidate match, and it takes a penalty value $W^d$ for dummy assignments which handle situations where a model point is not found in the scene:
\begin{equation}
U(z_i)=
\left \{
\begin{array}{ll}
W^d                                 & \textrm{if } z_i=\epsilon,\\
||f_i - f'_{z_i}||           &  \textrm{else,}\\
\end{array}
\right .
\label{eq:termu}
\end{equation}
$f_i$ and $f'_{z_i}$ being respectively the feature vector of model point i, and the feature vector of scene point $z_i$.

The $D$ term is based on angles. Since our data is embedded in space-time, angles include a temporal component not related to scale changes induced by zooming. We therefore split the geometry term $D$ into a temporal distortion term $D^t$ and a spatial geometric distortion term $D^g$, weighted by a parameter $\lambda_3$:
\begin{equation}
D(z_i,z_j,z_k) = D^t(z_i,z_j,z_k) + \lambda_3 D^g(z_i,z_j,z_k)
\label{eq:termd}
\end{equation}
where the temporal distortion $D^t$ is defined as time differences over two pairs of nodes of the triangle:
\begin{equation}
D^t(z_i,z_j,z_k) = \Delta(i,j) + \Delta(j,k)
\label{eq:timedistortiongeneral}
\end{equation}
with:
\begin{equation}
\Delta(i,j) = |(t(i)-t(j)) - (t'(z_i)-t'(z_j))| 
\label{eq:delta}
\end{equation}
Here, $\Delta(i,j)$ is the time distortion due to the assignment of model node pair $(i,j)$ to scene node pair $({z_i},{z_j})$. The temporal distortion term penalizes the discrepancy in the extent of time between model node pairs and the corresponding scene node pairs. The model node pairs should not be too close or too far from each other likewise the scene node pairs. Finally, $D^g$ is defined over differences of angles:
\begin{equation}
D^g(z_i,z_j,z_k) =
\left | \left |
\begin{array}{l}
a(i,j,k)-a'(z_i,z_j,z_k)  \\
a(j,i,k)-a'(z_j,z_i,z_k)  \\
\end{array}
\right | \right |.
\\
\end{equation}
Here, $a(i,j,k)$ and $a'(z_i,z_j,z_k)$ denote the angles subtended at point $j$ for, respectively, model triangle indexed by $(i,j,k)$ and scene triangle indexed by $(z_i,z_j,z_k)$. $\lVert . \rVert$ is the L2 norm. The difference between angles takes into account the circular domain of angles.
}

\subsection{Approximations}
\label{sec:approximations}

\noindent
In the context of activity recognition, the geometric data are embedded in space-time. We make the following assumptions relative to the temporal domain to derive an efficient minimization algorithm:

\hypot{1: Causality}
While objects (and humans) can undergo arbitrary geometrical transformations like translation and rotation, which is subsumed by geometrical matching invariance in our formulation, human actions can normally
 \emph{not} be reversed.
In a correct match, the temporal order of the points should be retained, which can be formalized as
follows
\begin{equation}
\forall \ i,j :
t(i) \le t(j) \Leftrightarrow t'(z_i) \le t'(z_j)
\label{eq:causality}
\end{equation}
where the notation $t(i)$ stands for the temporal coordinate of model interest point $i$, and $t'(z_i)$ stands for the temporal coordinate of scene interest point $z_i$.

\hypot{2: Temporal closeness}
Another reasonable assumption is that the extent of time warping between model and scene time axes must be limited. In other words, two points which are close in time must be close in both the model set and the scene set. Since our graph is created from proximity information (time distances have been thresholded to construct the hyperedges), this can be formalized as follows:
\begin{equation}
\forall \ i,j,k \in \mathcal{E}:
|t'(z_i) - t'(z_j)| < T \ \wedge \
|t'(z_j) - t'(z_k)| < T
\label{eq:temporalcloseness}
\end{equation}
where $T$ is a parameter.

We have shown in \cite{CeliktutanWolfMIV2013}, that this problem can be solved in polynomial time if the data are embedded in space-time, as opposed to the general class of  problems, which is NP-hard \cite{Torresani2008ECCV}.

Due to the sequential nature of activities, graphs obtained from data embedded in space-time are generally elongated in time, i.e. the graphical structure is heavily influenced by the temporal order of the nodes. This is particularly true in the case of graphs constructed from space time interest points, which are commonly very sparse. Typical interest point detectors extract only few such points per frame --- typically between $0$ and $5$ \cite{Laptev_hoghof_CVPR08}. We take advantage of this observation to restrict the model graph by keeping only a single interest point per model frame. This is done by choosing the most salient one, i.e., the one with the highest confidence of the interest point detector.

We also restrict the set $\mathcal{E}$ of model graph edges to connections of each model point $i$ to its two immediate frame-wise predecessors $i-1$ and $i-2$ as well as to its two immediate successors $i+1$ and $i+2$. This creates a planar graph with triangular structure, as illustrated in Figure \ref{fig:stipsgraph}b. A video is described by the way this planar graph twists in space time, as well as the appearance features associated with each node. According to the visual content of a video, there may be frames which do not contain any space time interest points, and therefore no nodes in the model graph. These empty frames are not taken into account, e.g. in Equation (\ref{eq:temporalcloseness}), when triplets of consecutive frame numbers are considered.

\subsection{Minimization}
\label{sec:minimization}
\noindent
The neighborhood system of this simplified graph can be described in a very simple way using the node indices of the graph, similar to the dependency graph of a second order Markov chain. The general energy given in (\ref{eq:globalenergyhypershort}) can therefore be expressed simpler as follows:
\begin{equation}
E(z) = \sum_{i=1}^M U(z_i) + \sum_{i=3}^M D(z_i,z_{i-1},z_{i-2})
\end{equation}
where we also have absorbed $\lambda_1$ and $\lambda_2$ into $U$ and $D$, respectively.
The elongated form of the graph allows us to derive an efficient inference algorithm for calculating 
$
\widehat{z} = \arg \min_z E(z)
$
based on the following recursion:
\begin{equation}
\alpha_i(z_{i-1},z_{i-2})
=  \displaystyle{\min_{z_i} }
\bigg [
U(z_i) + D(z_{i},z_{i-1},z_{i-2})  +  \
\alpha_{i+1}(z_i,z_{i-1}) \
\bigg ]
\label{eq:recursion2ndorder}
\end{equation}
with the initialization
\begin{equation}
\alpha_M(z_{M-1},z_{M-2}) = \min_{z_M}
\left [
U(z_M) + D(z_M,z_{M-1},z_{M-2})
\right ]
\label{eq:init}
\end{equation}
During the calculation of the trellis, the arguments of the minima in Equation (\ref{eq:recursion2ndorder}) are stored in a table $\beta_i(z_{i-1},z_{i-2})$. Once the trellis is completed, the optimal assignment can be calculated through classical backtracking:
\begin{equation}
\widehat{z_i} = \beta_{i}(z_{i-1},z_{i-2}),
\label{eq:backtracking}
\end{equation}
starting from an initial search for $z_1$ and $z_2$:
\begin{equation}
(\widehat{z_1},\widehat{z_2}) = \arg \min_{z_1,z_2} [U(z_1) + U(z_2) + \alpha_3(z_2,z_1)]
\label{eq:initsearch}
\end{equation}
The computational complexity and the memory complexity of the algorithm are defined by the trellis, a $M{\times}S{\times}S$ matrix, where each cell corresponds to a possible value of a given variable. The calculation of each cell requires to iterate over all $S$ possible combinations of $z_i$. It is easy to see that the computational complexity is $O(S^3M)$ and that the memory complexity is $O(S^2M)$.

Exploiting the different assumptions on the spatio-temporal data given above, the computational complexity can be decreased further. A large amount of combinations from the trellis can be pruned applying the following constraints: 
\begin{itemize}
\item given variable $z_i$, all values for its predecessors $z_{i-1}$ and $z_{i-2}$ must be necessarily \textit{before} $z_i$, i.e. lower.
\item given variable $z_i$, we will allow a maximum number of $T$ possibilities for the values of the successors $z_{i+1}$, $z_{i+2}$, which are required to be \textit{close}.
\end{itemize}
These pruning measures decrease the complexity to $O(SMT^2)$, where $T$ is a small constant measured in the number of frames. Since $T$ is a small constant, the computational complexity therefore becomes linear on the number of points in the scene: $O(SM)$. 

Let us note that no such prior pruning is applied to the scene frames, which therefore may contain an arbitrary number of points. At a first glimpse it could be suspected that the single-point-per-frame approach could be too limited to adequately capture the essence of an action sequence. Experiments have shown, however, that the single chain performs surprisingly well. It should be noted again, that no restrictions have been imposed on the scene, in other words, none of the scene points have been eliminated.


\section{A parallel solver for GPUs}
\label{sec:parallelsolver}

\begin{figure}[h!]
\centering
\def\svgwidth{\columnwidth}
\textsf{ 
\executeiffilenewer{memory_architecture.svg}{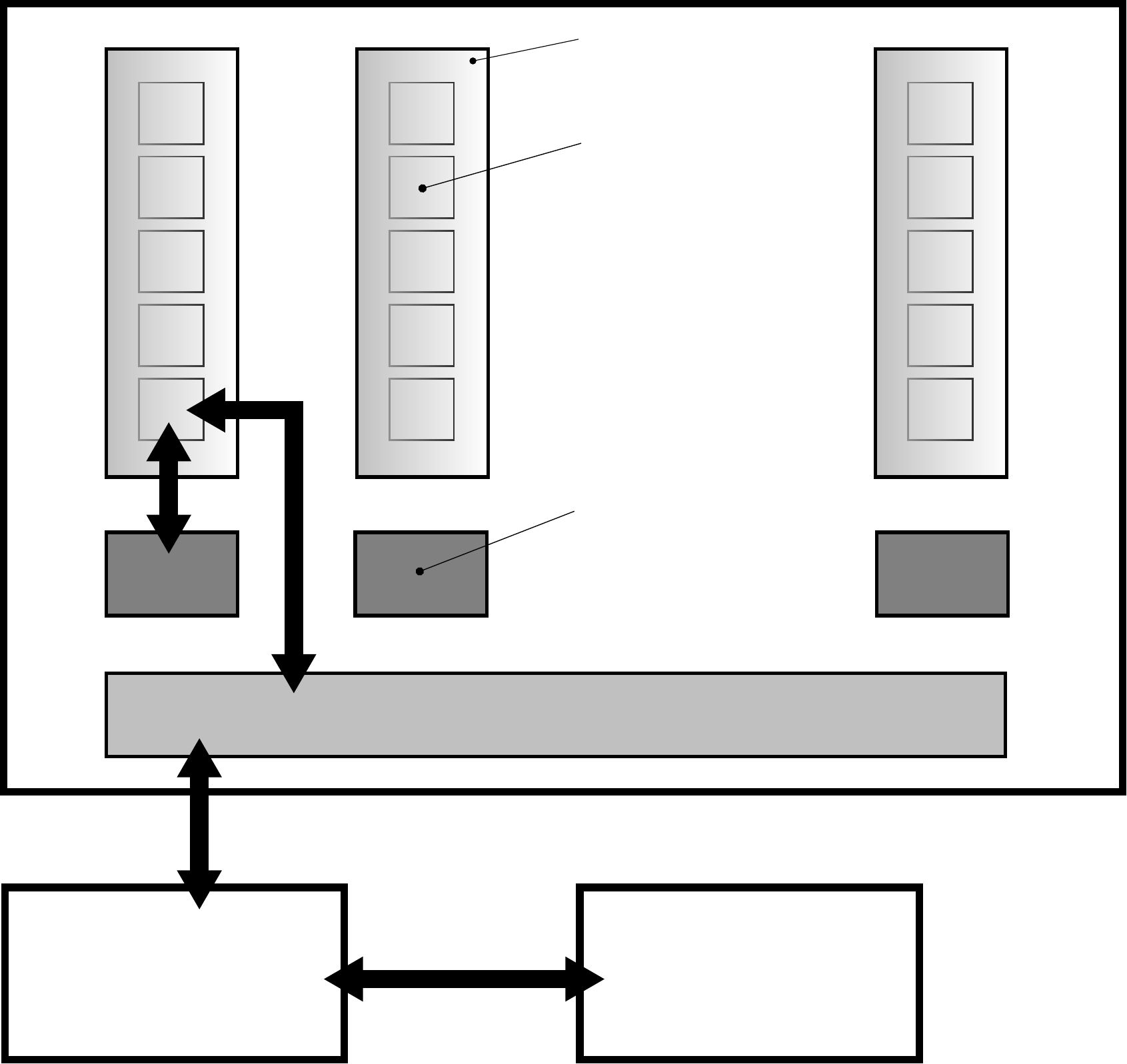}%
{inkscape -z -D --file=memory_architecture.svg --export-pdf=memory_architecture.pdf --export-latex ; mv memory_architecture.pdf_tex memory_architecture.pdf.in.tex}%
\input{memory_architecture.pdf.in.tex}%

}
\caption{The architecture of a CPU+GPU system as seen by the hardware independent OpenCL library.\label{fig:memoryarchitecture}}
\end{figure}

\noindent
Solving the matching problem requires computing Equations (\ref{eq:init}), (\ref{eq:recursion2ndorder}), (\ref{eq:backtracking}) and (\ref{eq:initsearch}). However, the computational and memory complexity are dominated by the requirement to solve (\ref{eq:recursion2ndorder}) for different indices $i$ and for different combinations of $z_{i-1}$ and $z_{i-2}$, which boils down to filling a three dimensional array 
with values, taking into account certain dependencies between these values. In the following section we will present a parallel solver for this problem designed for modern GPUs. Although the system has been implemented using the hardware independent OpenCL library, we kept the description as library independent as possible.
For this reason, and to make the paper self-contained, we first define some common terms (and refer to Figure \ref{fig:memoryarchitecture} for a block scheme of modern GPU architecture):
\begin{description}
  \item[Kernel] --- a kernel is a piece of code which is executed by the GPU in parallel on several independent hardware units;
  \item[Work-item] --- the execution of a kernel on a single data package is called a work-item;
  \item[Work-group] --- work-items can be grouped into work-groups; all work-items of one work-group share data in local memory;
  \item[Global memory] --- global GPU memory can be accessed by all work-items. It is slower than local GPU memory;
  \item[Local memory] --- local GPU memory is shared by the work-items of the same work-group. It is faster than global GPU memory;
  \item[RAM] --- the computer's (classical) main memory used by the CPU. It cannot be directly accessed by work-items, thus code running on the GPU.
\end{description}
Given the restrictions of memory access and execution on the GPU, parallel algorithms follow a classical three-step process:
\begin{itemize}
\item transfer data from RAM to global GPU memory;
\item eventually transfer data from global GPU memory to different local GPU memory banks;
\item execute the kernel multiple times in parallel on the GPU;
\item transfer results from GPU memory to RAM
\end{itemize}  
Dependencies between results may require more complex schemes and/or multiple iterations of this process.

\subsection{Defining the GPU kernel}
\begin{figure}[t]
\centering
\def\svgwidth{\columnwidth}
\textsf{
\executeiffilenewer{alphai_iterations.svg}{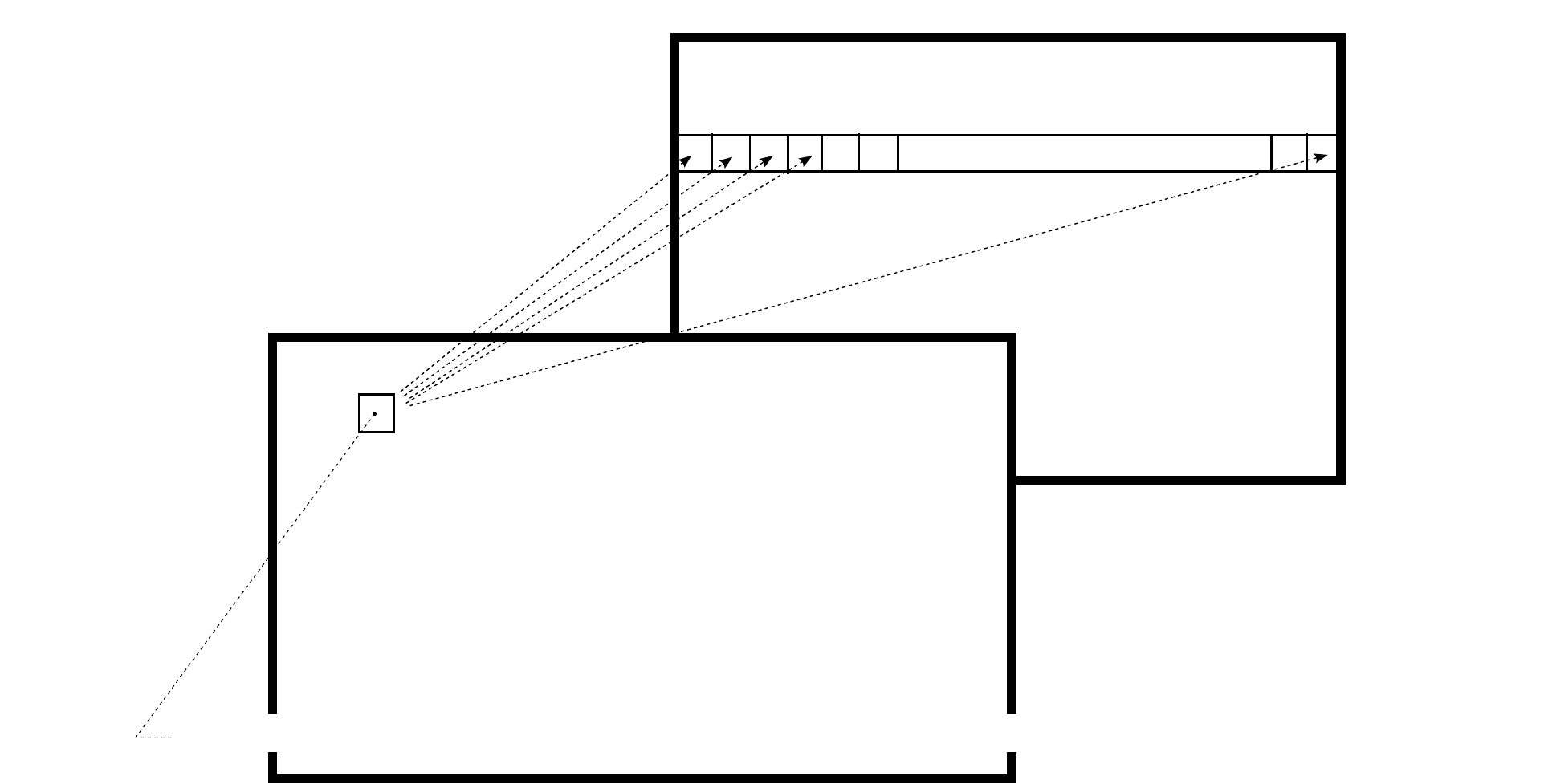}%
{inkscape -z -D --file=alphai_iterations.svg --export-pdf=alphai_iterations.pdf --export-latex ; mv alphai_iterations.pdf_tex alphai_iterations.pdf.in.tex}%
\input{alphai_iterations.pdf.in.tex}%

}
\caption{A visualization of the 3D trellis of Equation (\ref{eq:recursion2ndorder}) as a series of 2D tables of size $S{\times}S$. Each cell in table $\alpha_i$ is a result of a minimization operation taking as input results from a row in table $\alpha_{i+1}$.}
\label{fig:alphaiarray}
\end{figure}

\noindent
The recursion on $i$ in Equation (\ref{eq:recursion2ndorder}) produces (and works on) a 3D structure with 3 coordinates : ${i}$, $z_{i-1}$, $z_{i-2}$. Here, $i$ is the model frame/node index, which can also be interpreted as a temporal coordinate, and which takes values in $\{1..M\}$; $z_{i-1}$ and $z_{i-2}$ are assignment variables which can each take values in $\{1..S\}$. It is convenient to visualize this trellis as a series of 2D tables, as illustrated in Figure \ref{fig:alphaiarray}. The kernel code will execute the minimization in Equation (\ref{eq:recursion2ndorder}), where a single work-item deals with a triplet of parameters given by $i$, $z_{i-1}$ and $z_{i-2}$. Accordingly, a total number of $MS^2$ kernel executions (work-items) is required to fill the trellis.

The dependencies in Equation (\ref{eq:recursion2ndorder}) make it impossible to execute all kernels in parallel: it is easy to see, that a full row in $\alpha_{i+1}$ is required as input for each cell in $\alpha_i$. A scheduling is needed, which executes kernels in parallel with a block-wise sequential ordering, insuring that a whole $\alpha_{i+1}$ row is available, before the corresponding cell in $\alpha_i$ is executed. This scheduling could theoretically be done by the kernel itself through synchronization points. However, kernel-wise synchronization is only possible for work-items of the same work-group. The amount of work-items in the studied problem ($MS^2$) makes it unreasonable to use a single work-group for the whole problem. \change{Indeed, convenient values are M=30 and S=60. This corresponds to 1 seconds of model video and 2 seconds of scene video at 30 frames per second, using one node per frame. The amount of work items involved in the matching of the model graph to the scene graph, $MS^2$, is then 108000. This value is far above the capacity of even a high end GPU like the GTX580, where a workgoup is limited to 1024 work items. If we match a model against larger blocks of the scene, for instance entire videos, then this value will be even higher (see also table \ref{table:runtimes} for comparisons of two different scenarios with $S{=}754$ and $S{=}60$).}

A different solution is to perform synchronization through the CPU, i.e. launch parallel executions of a single array $\alpha_{i+1}$ on the GPU with the CPU taking over control between iterations over $i$. In other words, at the end of the kernel execution, the fall back to CPU execution acts as a synchronisation point for all work-items. Note, that the resulting scheduling is not different; a slight performance loss is due to the control flow change between the GPU and the CPU. This leads to a two-dimensional kernel, the dimensions being $z_{i-1}$ and $z_{i-2}$. A single loop in the kernel iterates over $z_i$ to perform the minimization operation in (\ref{eq:recursion2ndorder}).

\subsection{Pruning the trellis}
\label{seq:precomputingvalidity}

\begin{figure}[t] \centering
\input{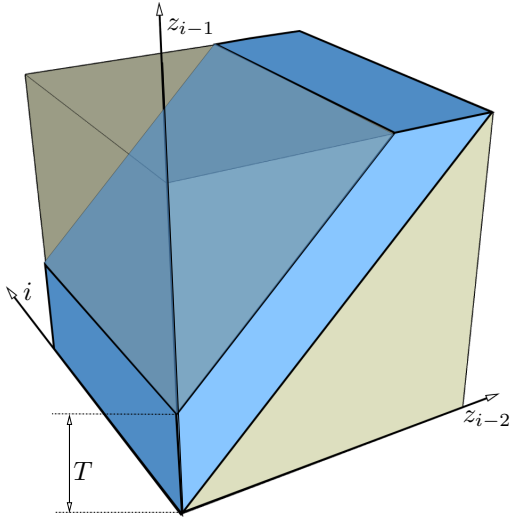}
\caption{The 3D trellis  of size $M{\times}{S^2}$ calculated by the recursive minimization algorithm. The shaded cross-section of size $\sim M{\times}S{\times}T$ in the middle  corresponds to the admissible combinations of $i$, $z_{i-1}$ and $z_{i-2}$.\label{fig:trellispruning}}
\end{figure}

\noindent
The causality assumption (\ref{eq:causality}) and the temporal closeness assumption (\ref{eq:temporalcloseness}) restrict the admissible combinations of $z_i$, $z_{i-1}$ and $z_{i-2}$, and therefore allow us to prune combinations in the trellis. In particular, for a given value of $z_{i-1}$, the value of $z_{i-2}$ is restricted to the interval $]z_{i-1} - T,z_{i-1}[$. This limits the admissible values of the trellis to a cross section of size $\sim S{\times}T$ in the 3D volume, as illustrated in Figure \ref{fig:trellispruning}. The validity tests for these constraints are precomputed and stored in a boolean array.
A similar reasoning restricts the values of $z_i$, c.f section \ref{seq:precomputingminloopboundaries}. 

\subsection{Precomputing the unary terms}
\label{sec:precomputingu}

\noindent
The complexity of the \change{unoptimized} kernel is dominated by the calculation of the unary terms $U(z_i)$, which are calculated in each iteration of the \change{minimization} loop, and which correspond to the Euclidean distance between the feature vector of a model node and the feature vector of a scene node. The size $F$ of the appearance feature vector may vary, common sizes are 50 to 200 components (162 in the case of our HoG/HoF features).

\change{
Assumptions (\ref{eq:causality}) and (\ref{eq:temporalcloseness}) decrease the part of $U$ for a work-item from $O(FS)$, to $O(FT)$, where ${T}{\le}S$ (see also sub section \ref{seq:precomputingminloopboundaries}). The unoptimized computational complexity for a whole array $\alpha_i$ is  $O(FST^2)$ and for the whole trellis it is $O(FSMT^2)$. 
Another way to derive the same result is to consider that each minimization in equation (\ref{eq:recursion2ndorder}) takes place over $T$ iterations, that the minimization is performed $SMT$ times (c.f. the blue cross section in figure \ref{fig:alphaiarray}), and that one term is of $O(F)$. 
}

\begin{figure}[t] \centering
\def\svgwidth{0.48\textwidth}
\textsf{ 
\executeiffilenewer{work_groups.svg}{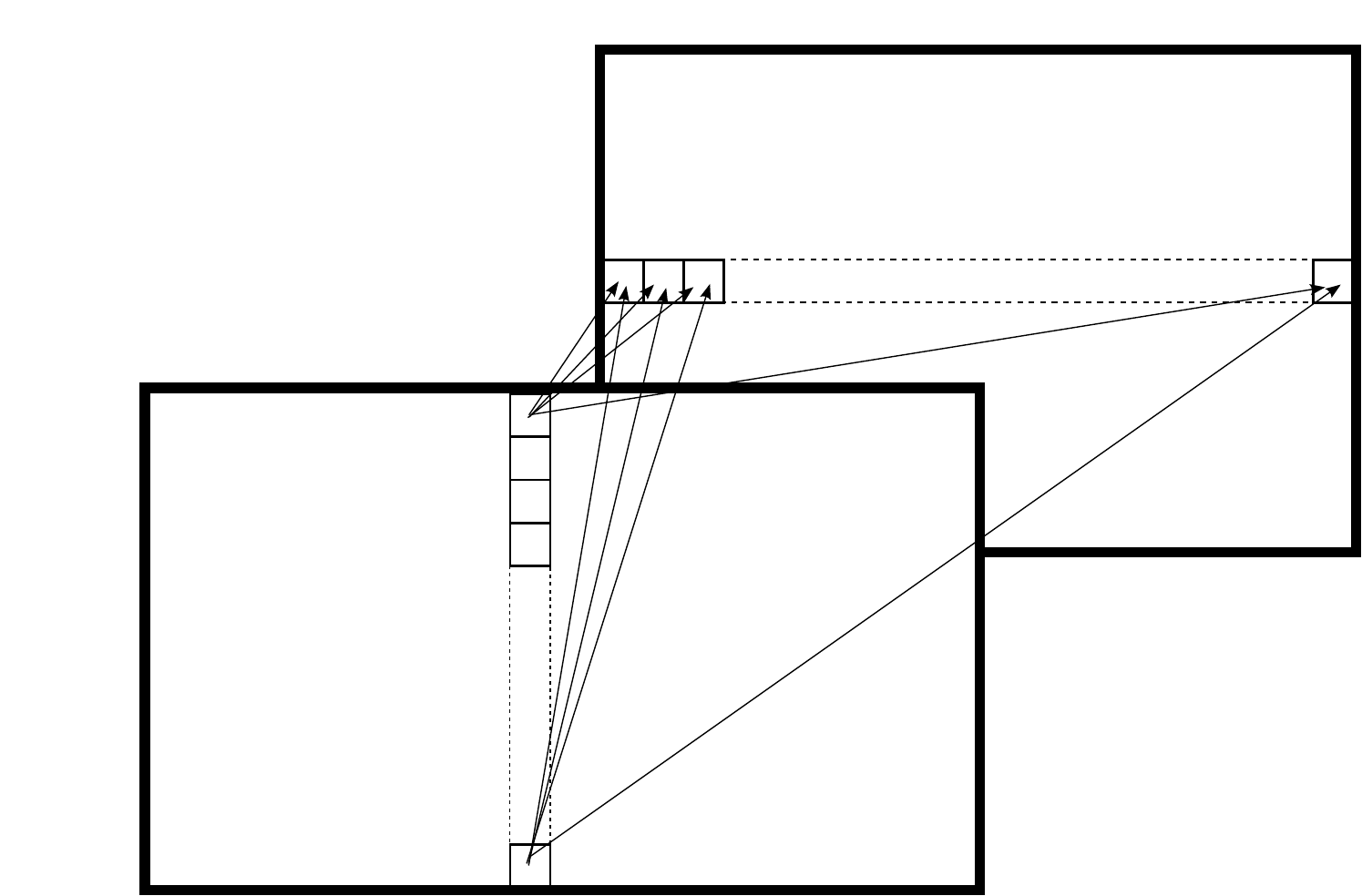}%
{inkscape -z -D --file=work_groups.svg --export-pdf=work_groups.pdf --export-latex ; mv work_groups.pdf_tex work_groups.pdf.in.tex}%
\input{work_groups.pdf.in.tex}%

}
\caption{Dependencies: one column of array $\alpha_i$ is handled by the work-items of the same work-group. The values of this column depend on a single row of previous array $\alpha_{i+1}$.}
\label{fig:workgroups}
\end{figure}

\change{
The unary terms take as input the model node and scene node. Therefore, out of the $SMT^2$ calls to $U$, only $SM$ different input arguments exist, which can be pre-calculated on the GPU by a separate kernel. The pre-computed unary terms are stored in a look-up table of size $S\times M$ in the GPU global memory. They are later used by the main GPU kernel, which calculates one $\alpha_i$ array. When the $\alpha_i$ kernel evaluates the matching of model node i to scene node $z_i$, it reads the corresponding pre-computed value of U in the look-up table.

Precomputing $U$ for all combinations of model nodes and scene nodes saves $T^2$ calls to $U$. The pre-computations themselves can be done in $O(FSM)$, so total computational complexity is reduced from 
$O(FSMT^2)$ to $O(SMT^2+FSM)$. For typical values of $T{=}10$ and $F{=}162$, the speed up factor is $\sim 61$.
}

\begin{figure}[t]
\centering
\def\svgwidth{\columnwidth}
\textsf{ 
\executeiffilenewer{distribution_of_scene_nodes.svg}{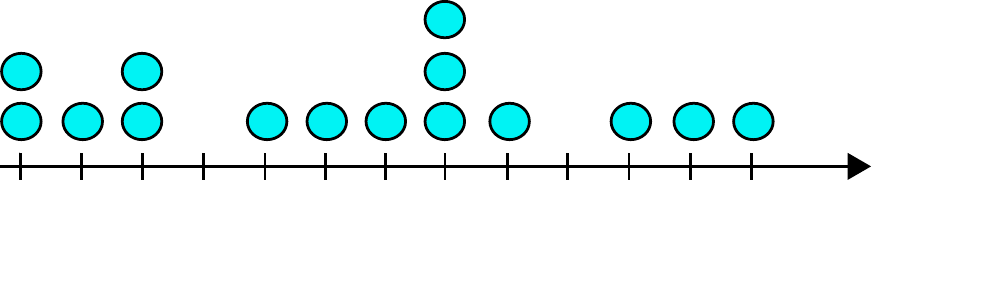}%
{inkscape -z -D --file=distribution_of_scene_nodes.svg --export-pdf=distribution_of_scene_nodes.pdf --export-latex ; mv distribution_of_scene_nodes.pdf_tex distribution_of_scene_nodes.pdf.in.tex}%
\input{distribution_of_scene_nodes.pdf.in.tex}%

}
\caption{\change{Example of a distribution of scene nodes over scene frames: as opposed to models, scenes may feature multiple nodes per frame. For example, frame 8 has nodes 9, 10 and 11.}\label{fig:distributionofscenenodes}}
\end{figure}

\change{
\subsection{Precomputing the minimization loop boundaries}
\label{seq:precomputingminloopboundaries}

\noindent
The minimization loop involved in equation (\ref{eq:recursion2ndorder}) is executed inside each kernel. During one kernel execution, i.e. for one work-item, $z_{i-1}$ and $z_{i-2}$ have a fixed value. The loop is then executed over $z_i$, whose value are limited by $z_{i-1}$ and $z_{i-2}$, according to causal constraint (\ref{eq:causality}) and temporal closeness constraint (\ref{eq:temporalcloseness}).
Applying the two constraints above and performing some algebraic simplifications results in two equations involving $z_i$:
\begin{equation}
\left \{
\begin{array}{l}
t'(z_i) \ge t'(z_i-1) \\
t'(z_i) < t'(z_i-2) + T \\
\end{array}
\right .
\label{eq:minloopboundaries2eq}
\end{equation}
Equations (\ref{eq:minloopboundaries2eq}) define an interval in temporal coordinates, which needs to be translated into an interval in scene nodes. Let us recall that the minimization in equation (\ref{eq:recursion2ndorder}) is over possible scene nodes. As opposed to model graphs, scenes can feature multiple nodes per frame, as illustrated in figure (\ref{fig:distributionofscenenodes}). This distribution of nodes over frames can be pre-calculated and stored in a table:
\begin{equation*}
minnode(f)=\inf(\{n: t'(n)=f\}), n=1...S
\end{equation*}
where $minnode(f)$ gives the first node of a given scene frame $f$, assuming that the scene nodes are sorted in temporal (i.e. frame) order. Then, the boundaries of the minimization loop in equation (\ref{eq:minloopboundaries2eq}) can be directly derived as follows:
\begin{equation*}
\left \{
\begin{array}{l}
\min(z_i){=} minnode(t'(z_{i-1})) \\
\\
\max(z_i){=} minnode(t'(z_{i-2}) + T) - 1 \\
\end{array}
\right .
\end{equation*}
where we took advantage of the fact that the maximum node of a frame $f$ is equal to $minnode(f+1)-1$.

Function $minnode(.)$ only depends on the distribution of the nodes in the scene. It can be pre-computed for each scene block, stored in GPU memory, and then shared by all work-items. 
}

\subsection{Work-groups and local GPU memory}
\label{seq:workgroupsandlocalgpumemory}

\noindent
Local GPU memory (NVidia ``shared memory'' or OpenCL ``local memory'') offers faster access than global GPU memory. Local memory is associated to a work-group, i.e., only work-items of this work-group can access it. Read and write operations to local memory can be synchronized between work-items of a single work-group. The goal is to optimize performance by performing a single transfer from global memory to local memory, followed by several accesses to local memory directly done by the kernels executions. Efficiency is guaranteed by organizing work-groups such that work-items can share common data. 

As illustrated in Figure \ref{fig:workgroups}, all cells of a single column of an array $\alpha_i$ depend on the same row of array $\alpha_{i+1}$. This can easily be derived from the recursive Equation (\ref{eq:recursion2ndorder}): calculating one column $\alpha_i(z_{i-1},z_{i-2})$ for fixed $i$ and $z_{i-1}$ requires the values $\alpha_{i+1}(z_i,z_{i-1})$ for fixed $i+1$ and $z_{i-1}$ over varying $z_i$, which corresponds to a row in the precedently calculated array. This naturally leads to a configuration where one column of the currently calculated array $\alpha_i$ is organized in a single work-group, where the column corresponds to the expression $\alpha_i(z_{i-1},z_{i-2})$ for fixed $i$ and $z_{i-1}$. 

The memory transfer of a whole row for a single workgroup is distributed over its work-items. As all tables $\alpha_{i}$ are of square shape (the number of columns equals the number of rows), a single value is transferred by each work-item at the beginning of the kernel execution. After a synchronisation point, the minimization routine itself is executed. This results in the kernel structure given in Algorithm \ref{alg:kernel}.

\begin{algorithm}[t]
\SetKwBlock{Kernel}{Kernel}{end}
\Kernel
{
	Copy a cell of $\alpha_{i+1}$ from global mem. to local mem.\;
	Wait for synchronization between all work-items\;
	Compute one cell of array $\alpha_i$ (minimization in (\ref{eq:recursion2ndorder}))\;
}
\caption{The kernel executed for each work-item. \label{alg:kernel}}
\end{algorithm}

\noindent
As a result, $S^2$ read operations from slow global memory per work-group have been replaced by $S$ read operations from global memory plus $S^2$ read operations from fast local memory. The total number of operations slightly grows by a small factor of $(S+S^2)/S^2$, which in practice is smaller than $1.017$ for the configuration we employed ($S{=}60$). However, the total access time decreases due to faster memory accesses.

\subsection{GPU memory persistence}
\label{seq:gpumemorypersistence}

\noindent
All model and scene data must be transfered from the CPU controlled RAM to the GPU, and hardware constraints force a transfer to global GPU memory. Let us recall that after a parallel computation of an array $\alpha_{i}$, execution falls back to the CPU, which controls the iterations over the different arrays $\alpha_{i}$ for different indices $i$. This could potentially involve a large number of memory transfers between CPU RAM and global GPU memory, as data needs to be exchanged between the CPU and the GPU before and after each iteration.

Data is persistent in global GPU memory between kernel executions, which can be exploited to minimize the number of memory transfers.
The full model data and the currently processed scene block are initially entirely transferred to the global GPU memory, and stay persistent between iterations over the tables $\alpha_i$. \change{The $\alpha_i$ arrays also stay in GPU memory between successive kernel executions.}

\subsection{Parallelizing calculations and transfer}
\label{seq:parallelizingcalculationsandtransfer}

\noindent
After calculation of the trellis, the optimal point assignment is calculated with the backtracking step given in Equation (\ref{eq:backtracking}). This step is of very low complexity (a single loop over $M$ iterations) and inherently sequential, therefore it is performed on the CPU. The whole trellis (tables $\alpha_i(z_{i-1},z_{i-2})$ for $i \in [1,M]$) must be transferred from global GPU memory to the CPU accessible RAM. Modern GPU architectures provide the ability to parallelize kernel executions and data transfer, which is here exploited to copy the results available in array $\alpha_{i+1}$ while kernel executions are computing the entries of $\alpha_i$. The CPU side of the matching algorithm is given in Algorithm \ref{alg:cpu}:

\begin{algorithm}[t]
\SetKwBlock{InParallel}{In parallel do}{end}
Copy full data from RAM to global GPU memory\;
\For{i=M to 1}
{
	\InParallel{
	Compute array $\alpha_i$ on GPU (run kernel)\;
	\If{$i{<}M$}
	{
		Copy array $\alpha_{i+1}$ from GPU to RAM\;
	}
	}
}
Copy array $\alpha_1$ from GPU to RAM\;
\caption{The CPU side algorithm.\label{alg:cpu}}
\end{algorithm}

\begin{table*}[t]
{\small
\begin{center}
\begin{tabularx}{\textwidth}{llllX}
\toprule
\bf{Work} &
\bf{Average perf.}	&				  		
\bf{Run-time}& 
\bf{Evaluation} &
\bf{Remarks on run-time}
\\

\midrule
Ta \etal \protect\cite{Ta_grapmatching_AVSS10}	& 
91.2\%				&
1.86s					& 
LOOCV 
&
s/frame, matching with 98 model graphs
\\
Borzeshi \etal	\cite{Borzeshi_PrototypeSelection_ICCVW11} & 
70.2\% 				& 
N/A						& 
Split 8/8/9 &
\\
Brendel \& Todorovic \cite{BrendelTodorovic2011}  	  					   		& 
N/A					& 
10s					& 
N/A &
Matching 1000 nodes graph with 2000+ nodes graph 
\\
Lv \& Nevatia \cite{LvNevatiaCVPR07}							&
N/A					& 
5.1s					&
N/A &
s/frame
\\ 
Savarese \etal \cite{Savarese_correlatons_WMVC08}							& 
86.8\%				& 
N/A						& 
LOOCV &
\\
Ryoo \& Aggarwal \cite{Ryoo_stmatch_ICCV09}								&
93.8\%				& 
N/A						& 
LOOCV &
\\
Mikolajczyk \& Uemura \cite{Mikolajczyk2011} 							&
95.3\%				& 
5.5s to 8s				& 
LOOCV & s/frame (-5s if SVM are not used)
\\ 
Baccouche \etal \cite{BaccoucheBMVC2012}&
95.8 &
N/A &
LOOCV & 
N/A 
\\
Jiang \etal	 \cite{JiangLinDavisPAMI12}							& 
93.4\%				&
N/A						&
LOOCV  \\     
\emph{\bf Our method on CPU}						&
91.0\%				& 
0.2s				& 
Split 8/8/9 
& s/frame, matching 754-nodes scene with 50 model graphs
\\
\emph{\bf Our method on GPU}						&
91.0\%				& 
0.02s				& 
Split 8/8/9 
& s/frame, matching 754-nodes scene with 50 model graphs
\\
\emph{\bf Our method on GPU}						&
91.0\%				& 
0.0035s				& 
Split 8/8/9 
& s/frame, matching 60-nodes scene with 50 model graphs
\\      
\bottomrule
\end{tabularx}
\end{center}
}
\caption{Comparison with the state-of-the-art methods on the KTH database (LOOCV = Leave-one-out-cross validation).\label{table:comparison}}
\end{table*}

\begin{table*}[t] \centering
\begin{tabularx}{\textwidth}{lrrrrrX}
\toprule
Implementation &
\#Scene &
\#Scene &
\multicolumn{2}{c}{Time/single model}&
\multicolumn{2}{c}{All 50 models} \\
&
nodes &
frames &
Tot(ms) &
per fr(ms) &
\multicolumn{2}{c}{Time/fr (ms)} \\
\toprule
CPU: Intel Core 2 Duo, &
754 &
723 &
2900	&
4.01 &
200.5
\\
E8600 @ 3.33 Ghz,
\\
Matlab/C(mex)
\\
\midrule
Nvidia GeForce GTS450, &
754 &
723 &
748 &
1.03 &
51.5
\\
192 cuda cores @ 1566 MHz, &
60 &
55 &
4 &
0.07 &
3.5 & real time!
\\
\change{mem bandwidth 21.3 GB/sec} &
\\
\midrule
Nvidia GeForce \change{GTX560}, &
754 &
723 &
405 &
0.56 &
28 & real time!
\\
336 cuda cores @ 1660 MHz, &
60 &
55 &
4 &
0.07 &
3.5 & real time!
\\
\change{mem bandwidth 128 GB/sec} &
\\
\midrule
\change{Nvidia GeForce GTX580,} &
754 &
723 &
\textbf{\color{blue}178} &
0.25 &
21 & real time!
\\
\change{512 cuda cores @ 1544 MHz,} &
60 &
55 &
4 &
0.07 &
3.5 & real time!
\\
\change{mem bandwidth 192 GB/sec} &
\\
\bottomrule
\end{tabularx}
\caption{Running times in milliseconds for two different GPUs and for two different scene block sizes. The last column on the right gives times per frame for matching the whole set of 50 model graphs. The bold blue value of \textbf{\color{blue}178 ms} is comparable to the values in table \ref{table:runtimeswithorwithoutoptimization}. \label{table:runtimes} }
\end{table*}

\begin{table*}[t] \centering
\begin{tabularx}{\textwidth}{llX}
\toprule
Implementation & Optimizations & Time (ms) \\
\toprule
CPU: Intel Core 2 Duo, &\ref{seq:precomputingvalidity}, \ref{sec:precomputingu} & 2900 \\
E8600 @ 3.33 Ghz, \\
Matlab/C(mex) \\
\midrule
Basic GPU Kernel & \ref{seq:precomputingvalidity}, \ref{sec:precomputingu} & 794 \\
Nvidia GeForce GTX580\\
\midrule
GPU Kernel with all alg. optimizations &
\ref{seq:precomputingvalidity}, \ref{sec:precomputingu} + \ref{seq:precomputingminloopboundaries}& 
326 \\
Nvidia GeForce GTX580 \\
\midrule
GPU Kernel with all alg. and archit. optimizations &
\ref{seq:precomputingvalidity}, \ref{sec:precomputingu}, \ref{seq:precomputingminloopboundaries}
+ 
& \textbf{\color{blue}178} \\
Nvidia GeForce GTX580  &
\ref{seq:workgroupsandlocalgpumemory}, \ref{seq:gpumemorypersistence}, \ref{seq:parallelizingcalculationsandtransfer}, \ref{seq:memorystructureoptimizations}
\\
\midrule
\midrule
GPU Kernel with all alg. and archit. optimizations &
\ref{seq:precomputingvalidity}, \ref{sec:precomputingu}, \ref{seq:precomputingminloopboundaries}
+ & 
1853 \\
Nvidia GeForce GTX580  & 
\ref{seq:workgroupsandlocalgpumemory}, \ref{seq:gpumemorypersistence}, \ref{seq:parallelizingcalculationsandtransfer}, \ref{seq:memorystructureoptimizations}
\\
&
(No closeness constraint: $T{=}\infty$) 
\\
\bottomrule
\end{tabularx}
\caption{\change{Contribution of different optimizations on execution time. Matching one model graph of 30 nodes to one scene graph of 754 nodes and 723 frames.}\label{table:runtimeswithorwithoutoptimization} }
\end{table*}

\subsection{Memory structure optimizations}
\label{seq:memorystructureoptimizations}

\noindent
An important concept in efficient computing on most GPU architectures is to ensure coalesced accesses to global GPU memory. Practically, this means that data in the global GPU memory has to be organised in contiguous blocks of multiples of 128 bytes to maximize access rate inside work-items. As a whole row of table $\alpha_{i+1}$ needs to be read for each cell (work-item) in $\alpha_i$, we store the different tables $\alpha_i$ in row order.

\change{
\subsection{Computational complexity and run-time}
\label{seq:complexityandruntime}

\noindent
As mentioned in section \ref{sec:minimization}, taking advantage of all approximations results in a computational complexity of $O(SMT^2)$, where $T$ is a small constant ($T{=}10$ in our experiments). However, that does not necessarily mean that run-time is a function of $O(SMT^2)$, as parallel processing is not yet taken into account. Actual run-time is of course considerably lower. If we consider $Q$ work-units (cores), run-time is of $O(\lceil \frac{ST}{Q}\rceil MT)$: $M$ kernel calls are launched, each one performing a minimization over $T$ iterations, and each kernel call scheduling $ST$ work-items. In the case where enough work-units are available on the GPU to perform all $ST$ work-items of one call in parallel, i.e. $Q\geq ST$, run-time is actually of $O(MT)$.

}

\section{Experimental Results}
\label{sec:experiments}

\noindent
The presented matching algorithm has been evaluated on a real-world application, namely action recognition in video sequences. The widely used public KTH database \cite{Schuldt_STJets_ICPR04} has been chosen as test dataset. It contains $25$ subjects performing $6$ actions (\textit{walking}, \textit{jogging}, \textit{running}, \textit{handwaving}, \textit{handclapping} and \textit{boxing}) recorded in four different scenarios including indoor/outdoor scenes and different camera viewpoints, totally $599$ video sequences (one is corrupted). The subdivision of the sequences is the same as the one provided on the official dataset website\footnote{http://www.nada.kth.se/cvap/actions/}. This results in 2391 subsequences in total. 

\change{The images of the KTH database are of size 160x120. However, only the preprocessing steps depend on the image size, the complexity of the matching algorithm itself is independent of it.}  

We use spatio-temporal interest points extracted with the 3D Harris detector~\cite{Laptev_hoghof_CVPR08} to constitute the nodes of the graph. Appearance features are the well-known HoG-HoF features extracted with the publicly available code in \cite{Laptev_hoghof_CVPR08}. 
The weighting parameters are set so that each distortion measure has the same range of values: $\lambda_1{=}0.6$, $\lambda_2{=}0.2$, $\lambda_3{=}5$, $T{=}10$ ($\lambda_3$ is explained in the appendix). 

All experiments use the leave-one-subject-out (LOSO) strategy. We augmented the number of model graph prototypes by selecting different sub-sequences and constructed each graph consisting of 20 to 30 frames containing at least one or more salient interest points. Action classes on the unseen subjects are recognized with a nearest prototype classifier (NPC). The distance between model and prototypes is based on the matching energy given in Equation (\ref{eq:globalenergyhypershort}). However, experiments showed that best performance is obtained if only the appearance terms $U(.)$ are used for distance calculation instead of the full energy (\ref{eq:globalenergyhypershort}). 

We learned a set of discriminative prototypes using sequential floating forward search (SFFS) \cite{SFBS}: model graph prototypes are created from the training subjects and the prototype selection is optimized over the validation set. 

The maximum admissible size of each work-group depends on the GPU itself and on the kernel specifically implemented for the problem. In our case, and for the Nvidia GeForce GTX 560 GPU, the limit is 768 work-items per work-group, which in practise is higher than the number of work-items we need: $S{=}60$ if scene blocks of 2 seconds are matched in video streaming mode, and $S{=}754$ if whole scene videos of 30 seconds are matched.

Floating point operations on Nvidia GPUs use different rounding algorithms than the FPUs on Intel CPUs, which may results in slightly different values obtained in the trellis. However, this did not impact the recognition results. 

We would like to point out that many results have been published on the KTH database, but most of the results cannot be compared due to different evaluation protocols, as has been studied on the detailed report on the KTH database in \cite{GoaChenHauptmann2010}. For completeness, we compare our performance with state-of-the-art methods. In Table~\ref{table:comparison}, we report average action recognition performance and computational time for the aforementioned methods in Section \ref{sec:introduction}. The results have been copied from the original papers. Although run-time calculations and used protocols differ between the papers, this table gives an overall idea that our proposed method is extremely competitive in terms of run-time with at the same time providing recognition performance comparable to the state of art. The best performing methods require several seconds of calculation per frame, whereas our method requires only 0.0035 seconds per frame. Our GPU algorithm is faster in several orders of magnitude.

\begin{table*}[t]
\caption{\change{Comparison of different model types with and without restriction to a single point per frame ($M$: number of model nodes, $S$: number of scene nodes,  $\overm$: number of model frames, $\overs$: number of scene frames, $R$: maximum number of interest points per frame in the \textit{scene} sequence, $N=3$: number of matched single-chain-single-point model).}}
\begin{center}
\begin{tabular}{llll}
\toprule
Method &Complexity & Accuracy &  \\
\midrule		
Proposed method (1 point per frame) & $O(SMT^2)$&	91.0\%\\
N=3 points per frame, method 1 (greedy)	& 	$O({\overs}{\overm}{T}^2R^N)$& 90.2\%	\\
N=3 points per frame, method 2 (independent chains)&	$O(SMT^2N)$ & 90.8\% \\  
\bottomrule
\end{tabular}
\end{center}
\label{table:differentmodels}
\end{table*}
 
The GPU implementation allows faster than real-time performance on standard medium end GPUs, e.g. a Nvidia GeForce GTS450. Table \ref{table:runtimes} compares run-times of the CPU implementation in Matlab/C (the critical sections were implemented in C) and the GPU implementation running on different GPUs with different characteristics, especially the number of calculation units. The run-times are given for matching a single model graph with 30 nodes against scene blocks of different lengths. If the scene video is cut into smaller blocks of 60 frames, which is necessary for continuous video processing, real time performance can be achieved even on the low end GPU model. With these smaller chunks of scene data, matching all 50 graph models to a block of 60 frames (roughly 2 seconds of video) takes roughly 3 ms regardless of the GPU model. 

\change{We can observe that the results for the three different GPUs (GTS450, GTX560 and GTX580) are identical when models are matched against scenes of 60 nodes. In this case, the number of cuda cores is higher than the amount of requested work items per call, i.e. per frame. Therefore, adding additional cuda cores does not give us an increase in performance. Other architectural differences (frequency, bandwidth) between the video cards are negligible and beyond the precision of time measurement. If we match one model against a whole video (754 frames), which requires much more work items, then the extra amount of cuda cores of the GTX560 and GTX580 cards makes a large difference in performance.}

The processing time of 3 ms per frame is very much lower than the limit for real time processing, which is 40 ms for video acquired at 25 frames per second. Additional processing will be required in order to treat overlapping blocks, which increases running time to 6 ms per frame.
\change{
The times given above also do not include interest point detection and feature extraction, but fast implementations of these steps do exist. As mentioned, in our experiments we used Laptev et al's STIP points \cite{Laptev_hoghof_CVPR08}, which performed slightly better than other detectors. Its pure CPU implementation requires 82 ms per frame. However, our own implementation of Dollar et al.'s points \cite{Dollar_STF_ICCV05} runs in 2 ms per frame including the detection of HoG features (also on CPU). These run-times can be decreased porting the algorithms to GPU, especially Dollar et al.’s detector, which essentially proceeds through linear filtering in time and space. The convolution operations can be easily performed on GPUs.
}

\change{In table \ref{table:runtimeswithorwithoutoptimization}, we compare execution times between the CPU implementation of the matching algorithm, and several implementations on GPU, with different types of optimizations. Comparing the CPU version running time (2900 ms) with the non-optimized GPU version (794 ms), we can see that the parallel hardware architecture of the GPU is almost 4 times faster than the CPU. The algorithmically optimized version of the GPU kernel (including optimizations described in sections \ref{seq:precomputingvalidity}, \ref{sec:precomputingu}, \ref{seq:precomputingminloopboundaries}) takes 326 ms to run, and is then 2.4 times faster than the non-optimized version. Finally, the GPU architectural optimizations described in sections \ref{seq:workgroupsandlocalgpumemory}, \ref{seq:gpumemorypersistence}, \ref{seq:parallelizingcalculationsandtransfer}, \ref{seq:memorystructureoptimizations} bring an additional speedup of factor 2 (178 ms vs. 326 ms). 
}

\begin{figure}[t] \centering
\subfloat[]{\includegraphics[height=6cm, angle=-90]{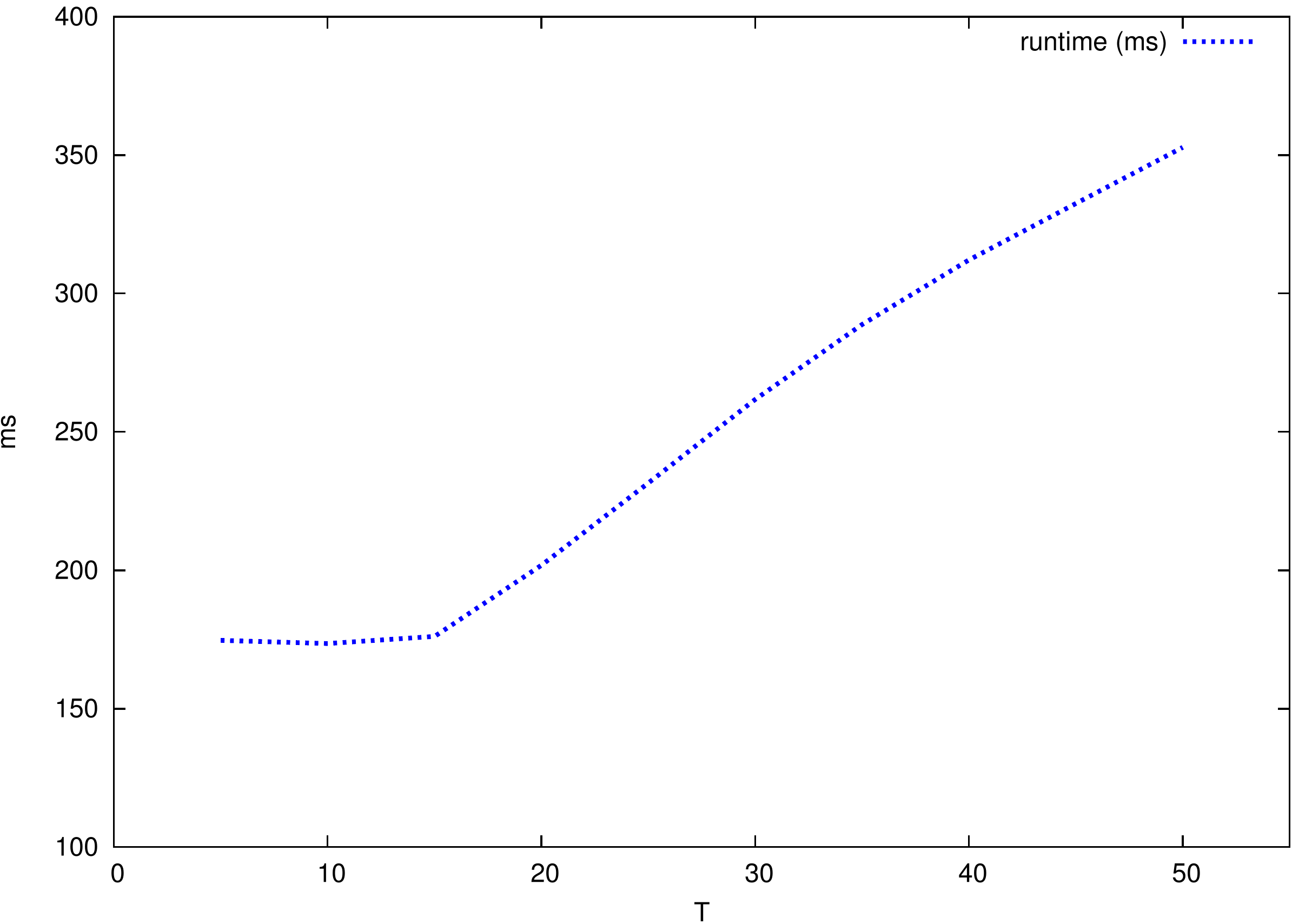}}\qquad
\subfloat[]{\includegraphics[height=6cm, angle=-90]{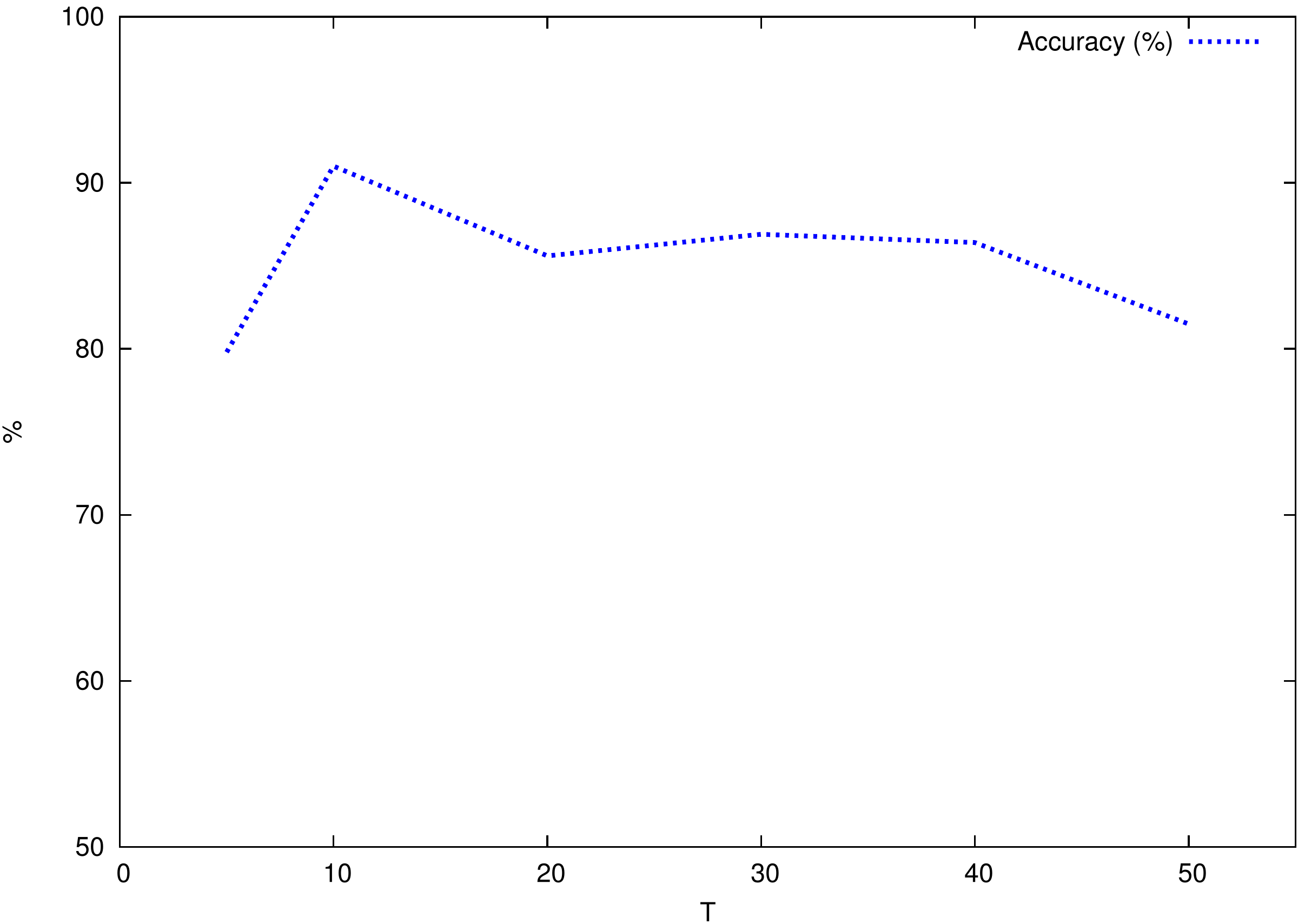}}\\
\caption{\change{Performance as a function of parameter $T$: (a) run-time; (b) classification performance (accuracy). No restrictions ($T=+\infty$) will lead to a runtime of 1853 ms (Nvidia GTX580, 754 scene nodes).\label{fig:curvet}}}
\label{3figs}
\end{figure}

\change{
The last line in the same table (table \ref{table:runtimeswithorwithoutoptimization}) shows a comparison with the fully optimized GPU version without temporal closeness constraint (1853 ms), giving the influence of the restriction of temporal warping.
Figure \ref{fig:curvet} shows this effect of parameter $T$ (restricting temporal warp) on performance. The parameter retained for all other experiments is $T{=}10$, and at this configuration we find the value of 178 ms also given in tables \ref{table:runtimes} and \ref{table:runtimeswithorwithoutoptimization}. Loosening the restrictions will not only significantly increase computational complexity, it will also decrease recognition performance. The restriction due to the parameter T allows to remove wrong assignments. No restrictions ($T=+\infty$) will lead to a runtime of 1853 ms.
}

\change{We also performed experiments to check the influence of the restriction to a single interest point per frame. The original formulation given in equation (\ref{eq:globalenergyhypershort}) is of polynomial complexity, but still too complex to solved in real time. We compared the proposed approximation to two different models allowing several interest points per frame :
\begin{description}
\item[Multiple points 1:] a greedy approximation of the the full model, where interest points assignments are decoupled from frame assignments. First, for each model frame and for all possible assigned scene frames, the optimal interest point assignments are made based on unary terms $U$ and inter-frame deformation terms $D$. Then, frames are assigned using the proposed model. 
\item[Multiple points 2:] creation of several single point models (several second order chains), each of which is solved independently. The resulting matching distance is given as the average over the different chains.
\end{description}
More details on these models are given in \cite{CeliktutanWolfMIV2013}. Results on the KTH are given in table \ref{table:differentmodels}. Results of the multiple point methods are comparable, even slightly lower. After investigating the results in detail, we believe that taking into account more points hurts performance because this decreases the invariance properties of the method. Space-time points are naturally very sparse, and taking more points actually leads to including less stable and less robust points.
}

\section{Conclusion}
\label{sec:conclusion}

\noindent
An efficient parallel algorithm for activity recognition has been presented, which resorts to parallel hypergraph matching on GPUs. The traditional problem of irregular computation flow in graph matching has been addressed by restricting the graph to a regular structure, which allows efficient inference by a recursive function working on a 3D trellis. The values in the trellis are computed in a block-wise parallel manner. The method is competitive with the state of the art in terms of recognition performance while at the same time being faster in several orders of magnitude. The current implementation is faster than real time on medium-end GPUs even though only a part of the computing units are currently used. Further speed-up could be gained by matching a scene video block against several model graphs in parallel and by distributing the cells of the multiple trellis over the computing units of the GPU.

\section*{Acknowledgement}

\noindent 
This work has been partially funded by the ANR project SoLStiCe (ANR-13-BS02-0002-01), a project of the grant program ``ANR blanc''.

{\small
\bibliographystyle{spmpsci}

\begin{thebibliography}{10}
\providecommand{\url}[1]{{#1}}
\providecommand{\urlprefix}{URL }
\expandafter\ifx\csname urlstyle\endcsname\relax
  \providecommand{\doi}[1]{DOI~\discretionary{}{}{}#1}\else
  \providecommand{\doi}{DOI~\discretionary{}{}{}\begingroup
  \urlstyle{rm}\Url}\fi

\bibitem{Aggarwal_review_ACM11}
Aggarwal, J.K., Ryoo, M.S.: Human activity analysis: a review.
\newblock ACM Computing Surveys  (2011)

\bibitem{Faggingerauer2012}
Auer, B.F., Bisselig, R.H.: A gpu algorithm for greedy graph matching.
\newblock In: Facing the Multicore Challenge II - Lecture Notes in Computer
  Science, pp. 108--119 (2012)

\bibitem{BaccoucheBMVC2012}
Baccouche, M., Mamalet, F., Wolf, C., Garcia, C., Baskurt, A.: Spatio-temporal
  convolutional sparse auto-encoder for sequence classification.
\newblock In: British Machine Vision Conference (BMVC) (2012)

\bibitem{Bergtholdt2010}
Bergtholdt, M., Kappes, J., Schmidt, S., Schn\"{o}rr, C.: {A Study of
  Parts-Based Object Class Detection Using Complete Graphs}.
\newblock International Journal of Computer Vision \textbf{87}(1-2), 93--117
  (2010)

\bibitem{Borzeshi_PrototypeSelection_ICCVW11}
Borzeshi, E.Z., Piccardi, M., Xu, R.Y.D.: A discriminative prototype selection
  approach for graph embedding in human action recognition.
\newblock In: ICCVW (2011)

\bibitem{BrendelTodorovic2011}
Brendel, W., Todorovic, S.: Learning spatiotemporal graphs of human activities.
\newblock In: ICCV (2011)

\bibitem{CaetanoCaelliBarone2006}
Caetano, T., Caelli, T., Schuurmans, D., Barone, D.: Graphical models and point
  pattern matching.
\newblock IEEE Tr. on PAMI \textbf{28}(10), 1646--1663 (2006)

\bibitem{CeliktutanWolfMIV2013}
Celiktutan, O., Wolf, C., Sankur, B., Lombardi, E.: Fast exact hyper-graph
  matching for spatio-temporal data.
\newblock Journal of Mathematical Imaging and Vision (to appear)  (2014)

\bibitem{Chen_MaxSubgraph_CVPR12}
Chen, C., Grauman, K.: Efficient activity detection with max-subgraph search.
\newblock In: CVPR (2012)

\bibitem{Dollar_STF_ICCV05}
Doll\'{a}r, P., Rabaud, V., Cottrell, G., Belongie, S.: Behavior recognition
  via sparse spatio-temporal features.
\newblock In: ICCV VS-PETS. Beijing, China (2005)

\bibitem{Conte_challengingcomplexity2007}
Donatello, C., Foggia, P., Vento, M.: Challenging complexity of maximum common
  subgraph detection algorithms: A performance analysis of three algorithms on
  a wide database of graphs.
\newblock Journal of Graph Algorithms and Applications \textbf{11}(1), 99--143
  (2007)

\bibitem{DuchenneCVPR09}
Duchenne, O., Bach, F.R., Kweon, I.S., Ponce, J.: A tensor-based algorithm for
  high-order graph matching.
\newblock In: CVPR, pp. 1980--1987 (2009)

\bibitem{DuchenneJoulinPonce2011}
Duchenne, O., Joulin, A., Ponce, J.: A graph-matching kernel for object
  categorization.
\newblock In: ICCV (2011)

\bibitem{FischlerRANSAC1981}
Fischler, M., Bolles, R.: Random sample consensus: A paradigm for model fitting
  with applications to image analysis and automated cartography.
\newblock Communications of ACM \textbf{24}(6), 381--395 (1981)

\bibitem{GoaChenHauptmann2010}
Gao, Z., Chen, M., Hauptmann, A., Cai, A.: Comparing evaluation protocols on
  the kth dataset.
\newblock In: Human Behavior Understanding, vol. LNCS 6219, pp. 88--100 (2010)

\bibitem{GareyJohnson1979}
Garey, M., Johnson, D.: Computers and Intractability: A Guide to the Theory of
  NP-Completeness.
\newblock W.H. Freeman (1979)

\bibitem{Gaur_StringFeatureGraphs_ICCV11}
Gaur, U., Zhu, Y., Song, B., Roy-Chowdhury, A.: A string of feature graphs
  model for recognition of complex activities in natural videos.
\newblock In: ICCV (2011)

\bibitem{Jenkins2011}
Jenkins, J., Arkatkar, I., Owens, J., Choudhary, A., Samatova, N.: Lessons
  learned from exploring the backtracking paradigm on the gpu.
\newblock In: International conference on Parallel processing, pp. 425--437
  (2011)

\bibitem{JiangLinDavisPAMI12}
Jiang, Z., Lin, Z., Davis, L.S.: Recognizing human actions by learning and
  matching shape-motion prototype trees.
\newblock PAMI  (2012)

\bibitem{Knight06thesinkhorn-knopp}
Knight, P.A.: The sinkhorn-knopp algorithm: Convergence and applications.
\newblock SIAM Journal on Matrix Analysis and Applications \textbf{30}(1),
  261--275 (2008)

\bibitem{Kollias2012}
Kollias, G., Sathe, M., Schenk, O., Grama, A.: Fast parallel algorithms for
  graph similarity and matching.
\newblock Tech. Rep. CSD-TR-12-010, Purdue University (2012)

\bibitem{Krizhevsky2012}
Krizhevsky, A., Sutskever, I., Hinton, G.E.: Imagenet classification with deep
  convolutional neural networks.
\newblock In: Neural Information Processing Systems (NIPS) (2012)

\bibitem{Laptev_hoghof_CVPR08}
Laptev, I., Marszalek, M., Schmid, C., Rozenfeld, B.: Learning realistic human
  actions from movies.
\newblock In: CVPR, pp. 1--8 (2008)

\bibitem{LauritzenSpiegelhalter1988}
Lauritzen, S., Spiegelhalter, D.: Local computations with probabilities on
  graphical structures and their application to expert systems.
\newblock Journal of the Royal Statistical Society B \textbf{50}, 157--224
  (1988)

\bibitem{LeeChoLee2011}
Lee, J., Cho, M., Lee, K.: Hyper-graph matching via reweighted random walks.
\newblock In: CVPR X (2011)

\bibitem{LeordeanuICCV2005}
Leordeanu, M., Hebert, M.: A spectral technique for correspondence problems
  using pairwise constraints.
\newblock In: ICCV, pp. 1482--1489. Washington, DC, USA (2005)

\bibitem{LeordeanuZanfirSminchisescu2011}
Leordeanu, M., Zanfir, A., Sminchisescu, C.: Semi-supervised learning and
  optimization for hypergraph matching.
\newblock In: ICCV 2011 (2011)

\bibitem{LiZhangLiuCSVT08}
Li, W., Zhang, Z., Liu, Z.: Expandable data-driven graphical modeling of human
  actions based on salient postures.
\newblock IEEE Tr. CSVT  (2008)

\bibitem{LinZengLiuZhu2009}
Lin, L., Zeng, K., Liu, X., Zhu, S.C.: Layered graph matching by composite
  cluster sampling with collaborative and competitive interactions.
\newblock CVPR \textbf{0}, 1351--1358 (2009)

\bibitem{LvNevatiaCVPR07}
Lv, F., Nevatia, R.: Single view human action recognition using key pose
  matching and viterbi path searching.
\newblock In: CVPR (2007)

\bibitem{Mikolajczyk2011}
Mikolajczyk, K., Uemura, H.: {Action recognition with appearance motion
  features and fast search trees}.
\newblock CVIU \textbf{115}(3), 426--438 (2011)

\bibitem{SFBS}
Pudil, P., Ferri, F.J., Novovicov, J., Kittler, J.: Floating search methods for
  feature selection with non-monotonic criterion functions.
\newblock In: ICPR, pp. 279--283 (1994)

\bibitem{RodenasFrencesc2011}
Rodenas, D., Serratosa, F., Sol-Ribalta, A.: Parallel graduated assignment
  algorithm for multiple graph matching based on a common labelling.
\newblock In: Graph-Based Representations in Pattern Recognition, \emph{Lecture
  Notes in Computer Science}, vol. 6658, pp. 132--141 (2011)

\bibitem{Ryoo_stmatch_ICCV09}
Ryoo, M.S., Aggarwal, J.K.: Spatio-temporal relationship match: video structure
  comparison for recognition of complex human activities.
\newblock In: ICCV (2009)

\bibitem{Savarese_correlatons_WMVC08}
Savarese, S., Delpozo, A., Niebles, J., Fei-Fei, L.: Spatial-temporal
  correlatons for unsupervised action classification.
\newblock In: WMVC. Los Alamitos, CA (2008)

\bibitem{Schuldt_STJets_ICPR04}
Schuldt, C., Laptev, I., Caputo, B.: Recognizing human actions: a local svm
  approach.
\newblock In: ICPR, pp. 32--36 (2004)

\bibitem{Shotton_depthpose_CVPR11}
Shotton, J., Fitzgibbon, A., Cook, M., Sharp, T., Finocchio, M., Moore, R.,
  Blake., A.: Real-time human pose recognition in parts from single depth
  images.
\newblock In: CVPR. Colorado Springs, USA (2011)

\bibitem{SinhaFrahm2006}
Sinha, N., Frahm, J.M., Pollefeys, M., Genc, Y.: Gpu-based video feature
  tracking and matching.
\newblock In: Workshop on Edge Computing Using New Commodity Architectures,
  vol. 278 (2006)

\bibitem{Ta_grapmatching_AVSS10}
Ta, A.P., Wolf, C., Lavoue, G., Ba\c{s}kurt, A.: Recognizing and localizing
  individual activities through graph matching.
\newblock In: AVSS (2010)

\bibitem{TodorovicECCV12}
Todorovic, S.: Human activities as stochastic kronecker graphs.
\newblock In: ECCV (2012)

\bibitem{Torresani2008ECCV}
Torresani, L., Kolmogorov, V., Rother, C.: Feature correspondence via graph
  matching: Models and global optimization.
\newblock In: ECCV, pp. 596--609 (2008)

\bibitem{VasconcelosRosenhahn2009}
Vasconcelos, C., Rosenhahn, B.: Bipartite graph matching computation on gpu.
\newblock In: International Conference on Energy Minimization Methods in
  Computer Vision and Pattern Recognition, pp. 42--55 (2009)

\bibitem{WangXiongYun2013}
Wang, G., Xiong, Y., Yun, J., Cavallaro, J.: Accelerating computer vision
  algorithms using opencl framework on the mobile gpu-a case study.
\newblock In: ICASSP, pp. 2629---2633 (2013)

\bibitem{ZaslavskiyBachVert2009}
Zaslavskiy, M., Bach, F., Vert, J.: A path following algorithm for the graph
  matching problem.
\newblock IEEE Tr. on PAMI \textbf{31}(12), 2227--2242 (2009)

\bibitem{ZassCVPR2008}
Zass, R., Shashua, A.: Probabilistic graph and hypergraph matching.
\newblock In: CVPR (2008)

\bibitem{ZhangZengJiITIP11}
Zhang, L., Zeng, Z., Ji, Q.: Probabilistic image modeling with an extended
  chain graph for human activity recognition and image segmentation.
\newblock ITIP  (2011)

\end{thebibliography}

}

\end{document}